\documentclass[letterpaper, 10 pt, conference]{./ieeeconf}

\IEEEoverridecommandlockouts                              % This command is only needed if 
\usepackage{graphics} % for pdf, bitmapped graphics files
\usepackage{epsfig} % for postscript graphics files
\usepackage{times} % assumes new font selection scheme installed
\usepackage{amsmath} % assumes amsmath package installed
\usepackage{amssymb}  % assumes amsmath package installed
\usepackage[ruled,vlined]{algorithm2e}
\usepackage{nomencl}
\usepackage[space]{cite}
\usepackage[linkcolor=black,citecolor=black,urlcolor=black,colorlinks=true]{hyperref}
\makenomenclature
\usepackage{tabu}
\usepackage{multirow}
\usepackage{subfigure}

\DeclareMathOperator*{\argmin}{arg\,min}

\DeclareGraphicsExtensions{.png,.jpg,.eps,.pdf}

\title{\LARGE \bf
Decentralized Visual-Inertial-UWB Fusion for Relative State Estimation of Aerial Swarm
}

\author{Hao Xu, Luqi Wang, Yichen Zhang, Kejie Qiu, Shaojie Shen
\thanks{All authors are with the Department of Electronic and Computer Engineering, Hong Kong University of Science and Technology, Hong Kong, China.
{\tt\small $\{$hxubc, lwangax, yzhangec, kqiuaa$\}$@connect.ust.hk, eeshaojie@ust.hk} \newline
This work was supported by HKUST-DJI Joint Innovation Laboratory and HKUST institutional fund.
}
}

\begin{document}

\maketitle
\thispagestyle{empty}
\pagestyle{empty}

%%%%%%%%%%%%%%%%%%%%%%%%%%%%%%%%%%%%%%%%%%%%%%%%%%%%%%%%%%%%%%%%%%%%%%%%%%%%%%%%
\begin{abstract}
The collaboration of unmanned aerial vehicles (UAVs) has become a popular research topic for its practicability in multiple scenarios. The collaboration of multiple UAVs, which is also known as aerial swarm is a highly complex system, which still lacks a state-of-art decentralized relative state estimation method. In this paper, we present a novel fully decentralized visual-inertial-UWB fusion framework for relative state estimation and demonstrate the practicability by performing extensive aerial swarm flight experiments. The comparison result with ground truth data from the motion capture system shows the centimeter-level precision which outperforms all the Ultra-WideBand (UWB) and even vision based method. The system is not limited by the field of view (FoV) of the camera or Global Positioning System (GPS), meanwhile on account of its estimation consistency, we believe that the proposed relative state estimation framework has the potential to be prevalently adopted by aerial swarm applications in different scenarios in multiple scales. 
% In recent years, aerial swarms have been used in real-world environments, including outdoor drone light shows, cooperative pesticide spraying, and inspection. These aerial swarms flying rely on global position method like GPS or motion capture system, which is not available in some complex application. In this paper, we propose a new framework for aerial swarm localization. With onboard sensors and onboard state estimations, we can estimate the real-time relative pose of every drone in the aerial swarm. This framework provides the ability of formation flying and cooperate working without any external devices. We evaluate the formation flying task base on our method, which could only finish with global localization methods. The framework can also apply to other movable robots.
\end{abstract}

\section{Introduction}\label{intro}

The collaboration of unmanned aerial vehicles (UAVs) has become a more and more popular research topic in the recent years for its practicability in inspection, search \& rescue, and light show performances. Aerial swarm is simply the collaboration of large amount of aerial vehicles. During the collaborative missions, it is crucial to perform real-time state estimation for each drone in order to plan smooth trajectories without collisions. Currently, the state estimations of aerial swarm are widely adopting either Global Positioning System (GPS) based methods\cite{jaimes2008approach}in outdoor environments or motion capture system and Ultra-WideBand (UWB) based methods\cite{alarifi2016ultra} \cite{ledergerber2015robot} \cite{mueller2015fusing} in indoor environments. However, the limitations of these methods are obvious: 
Due to the constraints of the GPS signal, state estimation for aerial swarm using GPS based methods can only be adopted in outdoor environments. Meanwhile, the system can only provide an accuracy at the magnitude of meters, which is not sufficient for tight formation flying. An alternative method is to increase the accuracy utilizing the Real-Time Kinematic positioning system\cite{moon2016outdoor}. Although the accuracy of centimeter-level positioning can be achieved by making use of the RTK-GPS framework, a ground station is compulsory to manipulate the system, causing the whole system to be clumsy and complicated to operate, which means that the framework can scarcely be adopted in an unknown environment. Despite that motion capture or traditional UWB based localization systems in\cite{alarifi2016ultra} \cite{ledergerber2015robot} \cite{mueller2015fusing} can be functional in the indoor scenarios where the GPS signal is not available, the installation external cameras or ground anchors are still required, which is not suitable for the navigation in unknown environments. 

\begin{figure}[t]
    \centering
    \includegraphics[width=1.0\linewidth]{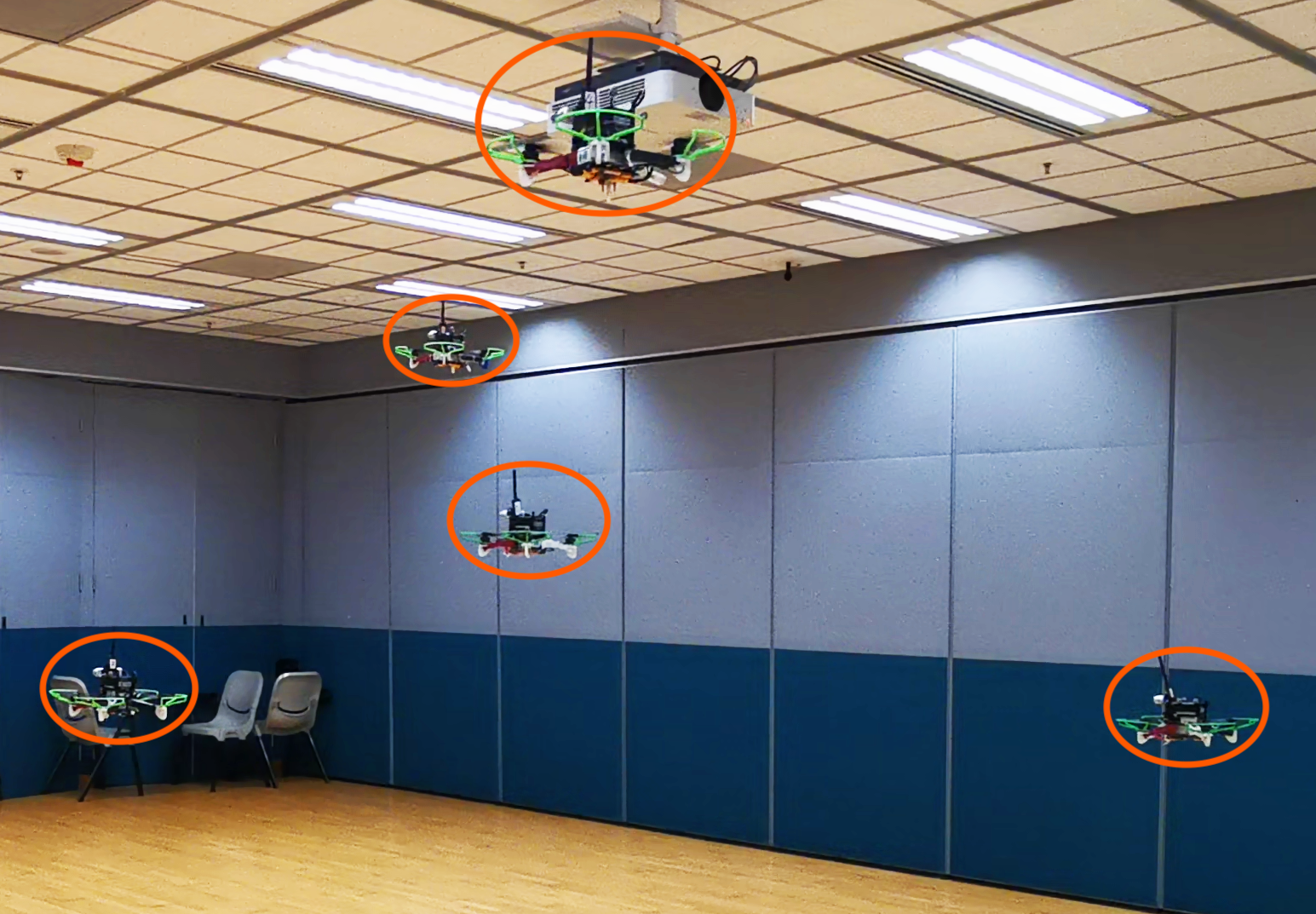}
    \vspace{-0.2cm}
    \caption{Indoor aerial swarm formation flight with 5 drones. The customized drone platforms are circled in the snapshot.}
    \label{fig:SWARM}
\vspace{-0.4cm}
\end{figure}

In order to get rid of the bulky external sensors, so as to have the capability of performing swarm flying in unknown environments, adopting a visual-inertial state estimation method\cite{qin2017vins} can be a possible solution, which can provide effective and robust state estimations for UAVs in indoor and outdoor, known and unknown environments. Nevertheless, visual-inertial odometry (VIO) is only capable of providing the state estimations in local frames. Together with the drift issue, it is apparent that a simple one-time alignment of the coordinate systems of the individual UAVs before take-off is not sufficient during formation flying. 
It is intuitive to perform detection for other vehicles from the images captured by the on-board cameras so as to obtain relative state estimations\cite{saska2017system}. Nonetheless, the estimation of the vehicles that are out of the field of view (FoV) of the on-board cameras can be a tricky problem. Hence, A sensor with omnidirectional sensing range and can be carried by a UAV, like a UWB module, is desired for the state estimation.

As a result, in this paper, we propose a vision-UWB based on-board real-time relative state estimation system which can be adopted in aerial swarm frameworks. The system is 
\begin{enumerate}
    \item Decentralized: Each vehicle perform relative state estimation individually and share information with others;
    \item Flexible: Only on-board sensors and computing power, including stereo cameras, inertial measurement units (IMU), UWB modules, as well as DJI Manifold2-Gs are utilized. No external off-board sensors or computing power are required;
    \item Robust: The system can handle the lack of the detection results as well as malfunctions of individual vehicles during swarm flights.
\end{enumerate}

Our approach presents a comprehensive solution for on-board relative state estimation, eliminating the environmental restrictions of previous localization methods, which provides other researchers an opportunity to apply aerial swarm motion planning techniques in complex environments.

\section{Related Works}\label{related_works}

In recent years, it is shown that researches on aerial swarms are more likely to focus on planning and control, and the relative state estimation is always leaped by adopting external positioning systems, including GPS\cite{jaimes2008approach}, motion capture system\cite{preiss2017crazyswarm} or UWB modules with anchers\cite{ledergerber2015robot}, which are all centralized system, requiring bulky external devices. There is still no state-of-art solutions for decentralized relative state estimations, which can be adopted on aerial swarm applications. Up to present, there are several proposed solutions or systems but the practicability or the performance are generally not satisfactory. As mentioned previously in Sec. \ref{intro}, it is intuitive to use sensors like camera and UWB modules to perceive relative state estimation and the recent researches generally follows the two branches.

Several works have adopted the UWB based methods for relative localization. In\cite{guo2017ultra}\cite{guo2019ultra}, the triangulation with distance measurements from the UWB modules are combined with the self displacement from optical flow method to obtain a relative state estimation. Only a 2-D translation is estimated and the initialization requires movement of a drone while other drones keep stationary, which limit the application scenario. In\cite{goel2017distributed}, another decentralized system using Extended Kalmen Filter to fuse the information from IMU and relative distance is introduced. The system is claimed to be the first attempt to realize a distributed localization framework with low cost sensors. Nevertheless, there are still large amount of challenges, including the precision problem preserved, making it difficult to put in practice.  

A series of work on vision-based relative state estimation has been conducted by M. Saska and his team. In\cite{saska2016auro}, a vision based framework is gradually developed. In the system, each drone is equipped with a circle marker for the detection purpose, which can be quite redundant. In \cite{walter2018icuas}\cite{walterRAL2019}, additional UltraViolet Direction And Ranging (UVDAR) sensor and active marker are installed on the drone for tracking purposes. Nonetheless, all the systems may encounter potential failure or drift when the markers are out of the FoV of the cameras, which limits the application scenario to be leader-follower formation.

\begin{figure*}[t]
    \centering
    \includegraphics[width=0.8\linewidth]{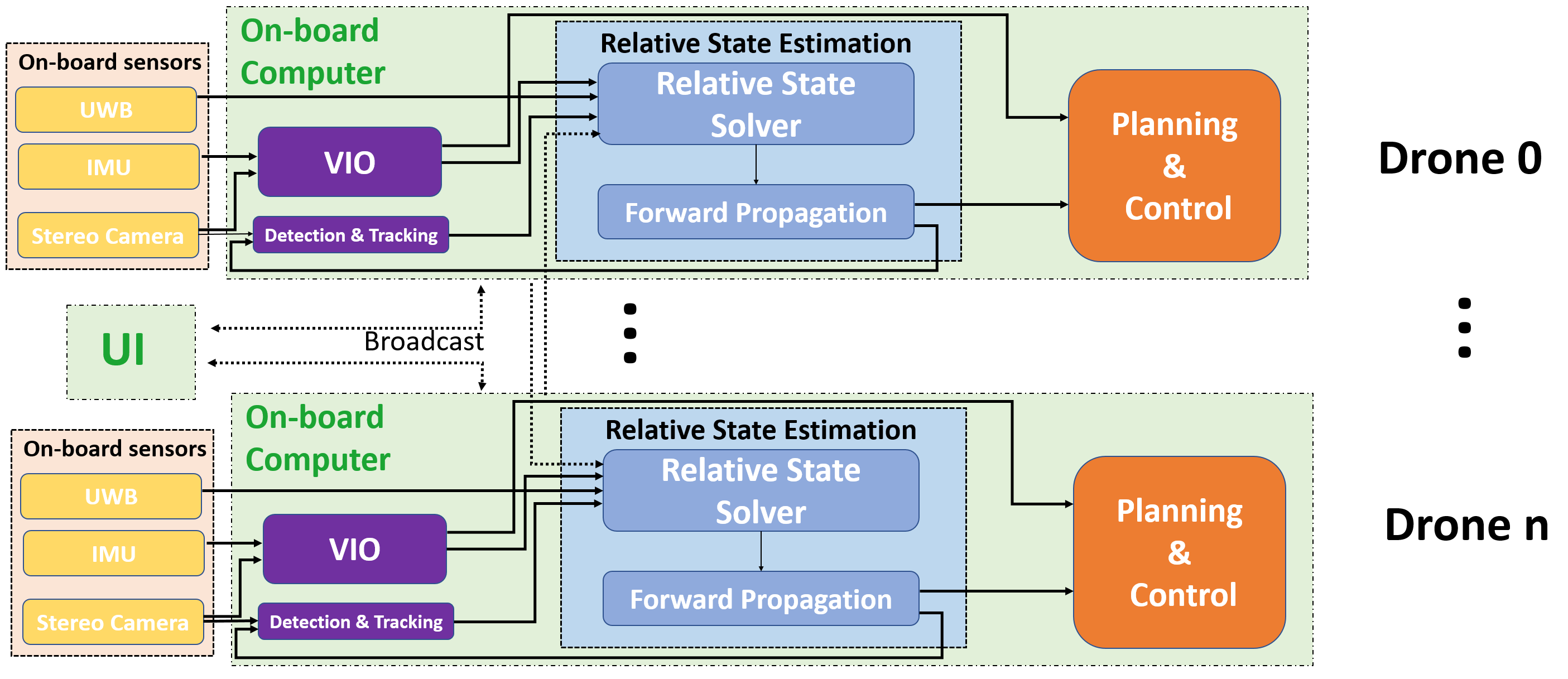}
    \vspace{-0.2cm}
    \caption{The system architecture of the whole aerial swarm framework. The data from the on-board sensors are processed, then broadcast to all the other terminals. The relative state estimator on each of the on-board computer collects both on-board and broadcast information, including the relative distance from the UWB modules, the VIO, as well as the detection results, and perform optimization and prediction to obtain real-time relative state estimations. The estimation result is sent back to facilitate the matching procedure of detection and tracking meanwhile serve the planning and control. The ground station obtain the real-time information from the drones to monitor the flight status and concurrently send the commands to the drones. All the communications between the devices are through UWB broadcast.}
    \vspace{-0.6cm}
    \label{fig:diagram}
\end{figure*}

% \begin{figure}
% % \centering
% \begin{subfigure}{.5\textwidth}
%     \centering
%     \includegraphics[width=\linewidth]{distance}
%     \caption{Distance between nodes}\label{ori_dis}
%     \end{subfigure}%
% \begin{subfigure}{.5\textwidth}
%         \centering
%         \includegraphics[width=\linewidth]{figure_3_to_7_error.png}
%         \caption{The distance error from node 3 to node 7}
%     \end{subfigure}
    
% \begin{subfigure}{.5\textwidth}
%         \centering
%         \includegraphics[width=\linewidth]{figure_3_to_8_error.png}
%         \subcaption{The distance error from node 3 to node 8}
%     \end{subfigure}%
% \begin{subfigure}{.5\textwidth}
%         \includegraphics[width=\linewidth]{figure_7_to_8_error.png}
%         \caption{The distance error from node 7 to node 8}
%     \end{subfigure}
%     \caption{Error of distance source data}    \label{errdis}
    
% Figure shows the distance measure by UWB in \ref{ori_dis} in meter, other figure shows the distance error between each two drones compared to motion capture system. Histogram plot shows that the distribution of the error is in  a wide range, some outliers can be found, which makes difficult to our method.
% \end{figure}

\section{Relative State Estimation Framework}\label{rel_est}

The complete relative state estimation structure is shown in Fig. \ref{fig:diagram}. A combination of a detector and a tracker is adopted for the motion capture of other drones when visible. Meanwhile, the measurements of relative distance from the UWB modules, together with the VIO are incorporated as well to optimize the 4 degrees of freedom (4DoF) relative state estimation, which consists of the three axes of translation $x$, $y$, $z$, and yaw rotation $\psi$.

We propose a two-stage optimization based relative state estimate for aerial swarm. The first stage is the fusion of visual and inertial data from stereo camera and IMU using VINS-Fusion\cite{qin2017vins}. The second stage is using the VIO result together with detection and UWB relative measurements for optimization-based fusion. The obtained relative state estimations are then sent back to facilitate the detection.
\subsection{Notations}\label{notation}
In order to help to understand the estimation framework, the following notations are defined below and followed by the rest of the paper.

\nomenclature[1]{$\hat{(\cdot)}$}{The estimated state.}

\nomenclature[2]{$ \mathbf{z}_{(\cdot)}^t $}{The measurement data at time t.}

\nomenclature[4]{$^{b_i}(\cdot)$}{State in drone \textit{i}'s body frame.}

\nomenclature[5]{$^{V_i}(\cdot)$}{State in drone \textit{i}'s VIO frame.}

\nomenclature[6]{$^{V_k}\mathbf{P}_{i}^t$}{Equals to $\begin{bmatrix}
\mathbf{R}_z( ^{V_k}\psi^t_{i}) & ^{V_k}\mathbf{x}_{i}^t \\
0  &  1
\end{bmatrix} $. The pose of drone \textit{i} in drone \textit{k}'s VIO frame at time $t$, which is from the result detecting and tracking. For simplification, the notation of $\mathbf{P}_{i}^t$ represents $^{V_i}\mathbf{P}_{i}^t$.$^{V_k}\mathbf{R}_z(^{V_k}\psi^t_{i})$ represent the rotation matrix rotate over z axis with angle $^{V_k}\psi^t_{i}$.}

\nomenclature[7]{$^{V_k}\mathbf{x}_{i}^t$}{Equals to $\left[^{V_k}x_{i}^t, { } ^{V_k}y_{i}^t, { }^{V_k}z_{i}^t\right]^T$. The translation part of $^{V_k}\mathbf{P}_{i}^t$.}

\nomenclature[8]{${\mathbf{\delta P}_{i}^t}$}{The transformation matrix from time $t-1$ to $t$ of drone \textit{i} from the VIO result, i.e. $\mathbf{ P}_{i}^t = {\mathbf{\delta P}_{i}^t} \mathbf{ P}_{i}^{t-1}$.}

\nomenclature[9]{$d_{ij}^t $}{Distance between drone \textit{i} and drone \textit{j} at time $t$.}

\printnomenclature

\subsection{Problem Formulation}

For an aerial swarm system that contains N drones, the relative state estimation problem can be represent as: For an arbitrary drone \textit{k}, estimate ${ }^{b_k} \mathbf{P}_i^t$ for each drone $\textit{i}$ at time $t$.

The formulation for relative state estimator is

\begin{equation}\label{relative_est}
  { }^{b_k}{\mathbf{\hat P}}_i^t = \argmin_{^{b_k}{\mathbf{\hat P}}_i^t} ({ }^{b_k}{\mathbf{P}}_i^{t})^{-1}({ }^{b_k}{\mathbf{\hat P}}_i^t).
\end{equation}

Together with the ego motion state $\mathbf{\hat P}_k^t$ estimated by the VIO, the pose of drone \textit{i} in drone \textit{k}'s VIO frame at time $t$ can be easily retrieved:
${ }^{V_k}\mathbf{\hat P}_i^t=\mathbf{\hat P}_k^t {}^{b_k}\mathbf{\hat P}_i^t$.

%The relative state estimate problem can be split into two parts,
%\begin{enumerate}
%    \item For a drone \textit{k} estimating the relative state of any other arbitrary drone \textit{i}, i.e. ${ }^{b_k} \mathbf{\hat P}_i^t$ 
%    \item Estimate the ego motion state of itself in a local frame, i.e. $\mathbf{\hat P}_k^t$.
%\end{enumerate}
%After solving these two problems, we can easily obtain arbitrary drone ${ }^{V_k}\mathbf{\hat P}_i^t=\mathbf{\hat P}_k^t {}^{b_k}\mathbf{\hat P}_i^t$.
%
%In this work, we focused on estimating  ${ }^{b_k}\mathbf{\hat P}_i^t$, i.e. the relative state estimation, and the ego motion parts is directly estimated by VIO. 

% The target of our approach is to provide a robust, distributed, and scalable method for swarm localizing problem.
\subsection{The First Stage: VIO}\label{vins}
As mention above, the first stage of the framework is to utilize the image from the stereo camera and the IMU data to obtain the VIO. In this stage, a GPU accelerated version of VINS-Fusion\footnote{\url{https://github.com/pjrambo/VINS-Fusion-gpu}}\cite{qin2019a}, an optimization based visual-inertial state estimator is adopted. The tightly coupled VIO result is treated as one of the ``measurements'' to the second stage. From the VIO of individual drones, the transformation matrix measurements at time t $\mathbf{z}_{\mathbf{\delta P}_{i}}^t$ can be obtained and formulated with a Gaussian model:

\begin{equation}\label{eq: delta_p}
    \mathbf{z}_{\mathbf{\delta P}_{i}}^t = {\mathbf{\delta P}_{i}}^t + \mathcal{N}(0,\,\sigma_{vio}^{2}) 
    %\vspace{-0.55cm}
\end{equation}

The equation does not contain the change of frame because ${\delta \mathbf{P}_{i}}^t$ is frame invariant.

Meanwhile, the VIO is also adopted for planning and control purposes of the system.

\subsection{Second Stage}
Besides the VIO result from the first stage, the relative measurements from detection and tracking as well as distance are required for the optimization in the second stage. 

\subsubsection{Detection and Tracking of Aerial Platforms}\label{detection_tracking}

For the detection of the aerial platforms, YOLOv3-tiny \cite{yolov3} is adopted, which is one of the state-of-art convolutional neural network (CNN) detectors, providing real-time object detection on the on-board computer. To utilized the detector for detecting our customized platforms, an additional training set captured by the on-board camera, as well as containing the drones are labeled and feed into the CNN. After the training, the network is able to efficaciously detect our aerial platforms. The detection results of 2-D bounding-boxs are combined with MOSSE trackers\cite{bolme2010visual} to provide the tracking results with a higher frequency.
Then, for each 2-D bounding-box from the tracking result, the depth of the center is retrieved from the depth camera. According to the pin-hole camera model, the 3-D relative position from the detection and tracking result can be recovered and the relative position in drone \textit{i}'s body frame $^{b_i}{\mathbf{z}_{DT}}^t_{j}$ can be retrieved according to the extrinsic parameters of the camera, where the footnote $(\cdot)_{DT}$ denotes the estimation result from detection and tracking. Besides the position of a bounding-box, the size also provides information about the depth. After the re-projection of a bounding-box onto the 3-D space, if the size immensely differs from the real size of the detected drone, the detection will be treated as an outlier. The detection and tracking result is also formulated with a Gaussian model:
\begin{equation}\label{eq: detection}
     ^{b_i}{\mathbf{z}_{DT}}^t_{j}=\left( (^{V_k}\mathbf{P}_{i}^{t})^{-1} { }^{V_k}\mathbf{P}_{j}^t \right)_T+ \mathcal{N}(0,\,\sigma_{det}^{2}) 
\end{equation}
%\begin{equation}
%D\begin{bmatrix} 
%u \\
%v \\
%1
%\end{bmatrix} = 
%\begin{bmatrix}
%f_x & 0   & c_x \\
%0   & f_y & c_y \\
%0   &   0 &  1
%\end{bmatrix} { ^{c_i}\mathbf{z}_{E_j}},    
%\end{equation}
%where $u, v$ denote the 2-D position of the center of a bounding-box on the image, $f_x, c_x, f_y, c_y$ are the intrinsic parameters of the camera, and $ ^{c_i}{\mathbf{z}_E}_{j}$ denotes the 3-D position of the detected drone \textit{i} in drone \textit{k}'s camera frame. T
where  $(\cdot)_T$ represents the translation in the transformation.

However, in our aerial swarm,  there is another issue that all the platforms have a similar appearance, which cannot be distinguished by the detector and the tracker. Therefore, a matching algorithm is required for labeling the drones with their unique IDs. Once a drone is detected, the position from the visual measurement and the final estimation of the whole framework are compared and the matches are performed based on the principle of least disparities meanwhile should be less than a preset threshold value. Otherwise, the measurement will be treated as an outlier as well. After the drones are successfully labeled, once a new detection frame arrives, if the bounding-boxs of the tracker and the detector with the same ID have sufficient overlap, both the detection and the tracking are valid. Otherwise the matching procedure will be re-executed.

\subsubsection{UWB distance measurements}\label{UWB_dist}
From the UWB modules, the relative distances between each pair of the UWB nodes can be obtained, which are the UWB distance measurements $\mathbf{z}_{d_{ij}}^t $. The measurements are also modeled with a Gaussian noise:
\begin{equation}\label{eq: distance}
    \mathbf{z}_{{d}_{ij}}^t =\left \Vert  ^{V_k}\mathbf{x}_{i}^t -{ } ^{V_k}\mathbf{x}_{j}^t \right \Vert_2 + \mathcal{N}(0,\,\sigma_{d}^{2}) 
    %\vspace{-0.55cm}
\end{equation}

%\subsection{Measurement Processing}\label{measurement}
%
%As the system diagram shown in Fig. \ref{fig:diagram}, the raw measurements are retrieved from 3 type of sensors:UWB module, Inertial Measurement Unit and Stereo camera. The measurements are first processed into the following three terms before optimization:
%
%
%
%The proposed relative state estimator is loosely coupled since the optimization is performed on the three terms above, which are not all raw measurements.  The three results are formulated with Gaussian noises compared with the ground truth data:
%\vspace{-0.4cm}

% It should be noted that all the processed measurements are shared by all the drones, and the equations above show the measurement models in any drone's frame. The three ``measurements'' are corresponded with the blue arrows, the black arrows and the red arrow in Fig. \ref{fig:methodology}. The covariances $\sigma_{d}^{2}$, $\sigma_{vio}^{2}$ and $\sigma_{det}^{2}$ are measured according to the ground truth data to make the model more precise.

\begin{figure*}[t]
    \centering
    \includegraphics[width=0.8\linewidth]{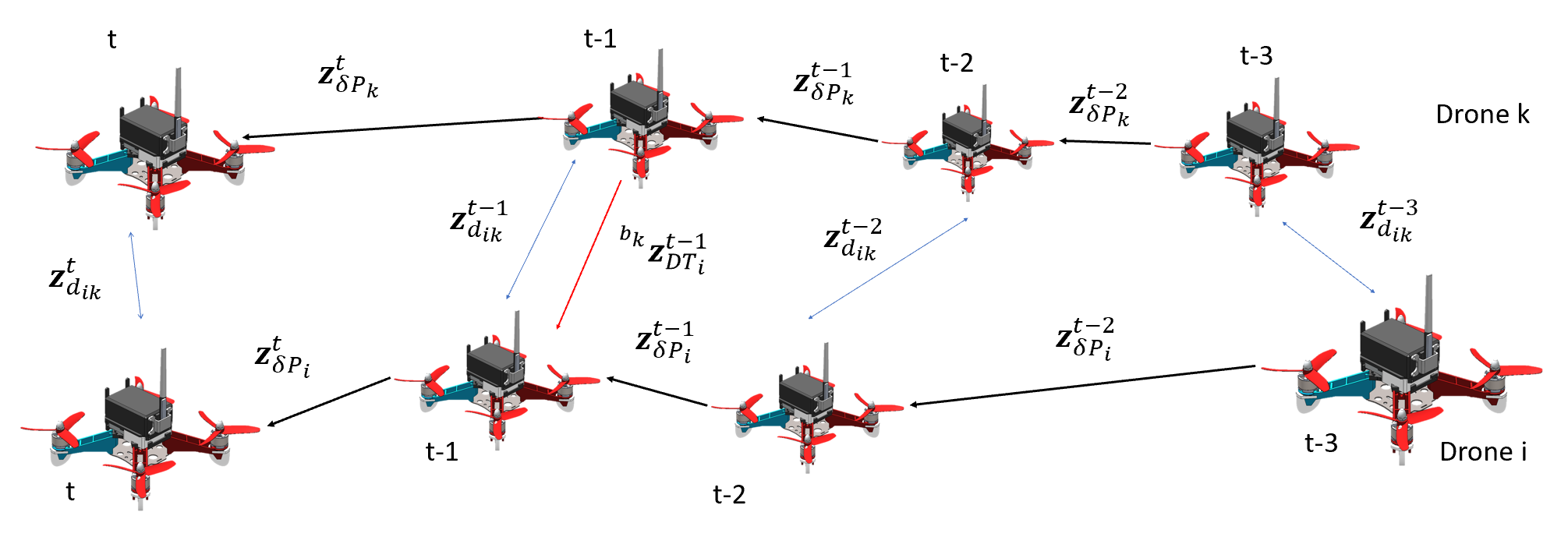}
    \caption{An illustration of the sliding window relative state estimation. The notations are stated in Sec. \ref{notation}. The black arrows represent the transformation between the poses, which are provided VIO; the blue arrows indicate the relative distance measurements from the UWB modules; the red arrow depicts the relative 3-D translation from the result of detection and tracking.}
    \vspace{-0.7cm}
    \label{fig:methodology}
\end{figure*}

\subsubsection{Data Frame, Keyframe Selection and Sliding Window}\label{KFs}
All the above mentioned measurements and VIO are broadcast to all the terminals and a frame consists of the measurements and VIO at a specific time:
\begin{itemize}
    \item The distance measurements of each two pair of drones $\mathbf{z}_{d_{ij}}^t $, provided by the UWB modules.
    \item The tightly-coupled VIO results $^{V_i}\mathbf{z}_{\mathbf{P}_{i}}^t$ from the ego state estimation of individual aerial platforms using VINS-Fusion\cite{qin2019a}, which is treated as ``VIO measurements''.
    \item The relative translation measurements obtained from the detector and tracker $^{b_i}{\mathbf{z}_{DT}}^t_{j}$  stated in Sec.\ref{detection_tracking}.
\end{itemize}

For drone \textit{k}, a frame is selected to be a keyframe if one of the following criteria is satisfied:
\begin{itemize}
    \item The frame contain new drones.
    \item Drone \textit{k} moves for a distance which is more than a certain threshold $td_0$ compare to last keyframe.
    \item Drone \textit{k} moves for a distance which is more than a certain threshold $td_1$ compare to last keyframe and it detects or is detected by other drones.
\end{itemize}
Where $td_0$ and $td_1$ are adjustable parameters and typically set following the relation of $0 < td_1 < td_0$. A sequence of keyframes in chronological order forms a sliding window for optimization. An illustration of a sliding window is shown in Fig. \ref{fig:methodology}. In practice, the maximum size of the sliding window for optimization is selected to be 50 to guarantee real-time and robust performance.

\subsubsection{Optimization-based Relative State Estimation}\label{optimization}
With the sliding window formed previously in Sec. \ref{KFs},  the optimization for the relative state estimation, which is the second stage can be performed. For a drone $k$, the full state vector $\mathcal{X}_k$ for optimization is defined as:
% \vspace{-0.2cm}
\[
\left[
\underbrace{\overbrace{^{V_k}\mathbf{\hat P}_{0}^{t_i T} \ ^{V_k}\mathbf{\hat P}_{0}^{t_{i + 1}T} ...\ ^{V_k}\mathbf{\hat P}_{0}^{t_{i + m - 1}T}}^{\text{m elements}} \ ^{V_k}\mathbf{\hat P}_{1}^{t_i T} ... \ ^{V_k}\mathbf{\hat P}_{n-1}^{t_{i + m - 1}T} }_{\text{m $\times$ n elements}}
\right]^T,
\]
where n is the amount of drones in the swarm system, m is number of keyframes in the sliding window.

Instead of directly solving Eq.\ref{relative_est}, a bundle adjustment formulation is adopted for the optimization of the relative states. By minimizing the Mahalanobis norm of the residuals of the measurements, a maximum a posterior estimation can be obtained.

The optimization is expressed as the following formulation:
\begin{equation}\label{opti_eq}
\begin{aligned}
    \min_{\mathcal{X}_k} \Bigg\{
    &\sum_{(i,j,t)\in\mathcal{L}}
    \left\Vert  \mathbf{z}_{d_{ij}}^t - \left\Vert  ^{V_k}\mathbf{\hat x}_{i}^t -  ^{V_k}\mathbf{\hat x}_{j}^t \right\Vert_2 \right\Vert^2 \\ + 
    &\sum_{(i,j,t)\in\mathcal{D}} \left\Vert \left( (^{V_k}\mathbf{\hat P}_{i}^{t})^{-1} { }^{V_k}\mathbf{\hat P}_{j}^t \right)_T - ^{b_i}{\mathbf{z}_{DT}}^t_{j}\right\Vert^2 \\ +
    &\sum_{(i,t)\in\mathcal{S}} \left\Vert \left(\mathbf{z}_{\mathbf{\delta P_{i}}}^t \right)^{-1} \left( ( ^{V_k}\mathbf{\hat P}_{i}^{t-1})^{-1} { }^{V_k}\mathbf{\hat P}_{i}^t\right) \right\Vert^2 \Bigg\}
\end{aligned}
% \vspace{-0.2cm}
\end{equation}
where  $\mathcal{L}$ is the set of all distance measurements, $\mathcal{D}$ is the set of all detection measurements, $\mathcal{S}$ is the set of all VIO ego motion measurements. $\left\Vert  \mathbf{z}_{d_{ij}}^t - \left\Vert  ^{V_k}\mathbf{\hat x}_{i}^t -  ^{V_k}\mathbf{\hat x}_{j}^t \right\Vert_2 \right\Vert^2$ is the residual of distance measurements;
$\left\Vert \left( (^{V_k}\mathbf{\hat P}_{i}^{t})^{-1} { }^{V_k}\mathbf{\hat P}_{j}^t \right)_T - ^{b_i}{\mathbf{z}_{DT}}^t_{j} \right\Vert^2$ is the residual of detection and tracking measurements, $(i,j)$ present the pair of successful detection and tracking of drone $j$ detected by drone $i$;
$\left\Vert \left(\mathbf{z}_{\mathbf{\delta P_{i}}}^t \right)^{-1} \left( ( ^{V_k}\mathbf{\hat P}_{i}^{t-1})^{-1} { }^{V_k}\mathbf{\hat P}_{i}^t\right) \right\Vert^2$ represents the residual of VIO results, ensuring the local consistency of the drone $i$'s trajectory measured in the two drones' VIO frames.

Simply solving Eq. \ref{opti_eq} can directly provide the result of the relative state estimation with:
$${ }^{b_k}\mathbf{\hat P}_{i}^t =(\mathbf{\hat P}_{k}^t)^{-1} {}^{V_k}\mathbf{\hat P}_{i}^t$$

The Ceres-solver\cite{ceres-solver}, which is an open-source C++ library developed by Google for modeling and solving the Non-linear Least-Squres optimization problems is adopted, while the Trust Region with sparse normal Cholesky method is chosen.
Meanwhile, since the planner or controller may require high frequency real-time poses, the real-time VIO from all the drones are composed with the optimized relative poses to provide a prediction at a higher rate of 100hz.

\subsubsection{Residual and Variable Pruning}
For an aerial swarm system containing $n$ drones utilizing $m$ keyframes in the sliding window, the optimization has $4mn$ variables to solve. In practice, five drones and 100 keyframes are adopted in our experiment, resulting in solving 2000 unknowns, which is time-consuming and unnecessary. In the aerial swarm scenario, hovering of several vehicles can be a usual case. In this situation, two residuals and the related variables can be pruned:

\begin{itemize}
    \item If drone \textit{i}, \textit{j} are hovering, then for any other drone \textit{k}, the distance residual mentioned previously in Sec. \ref{optimization} can be pruned, since the optimization is a formation of triangulation and the noise of distance measurements may cause the over-fitting issue.
    \item When a drone is hovering, the pose measurements should be fixed, meaning that the VIO residuals vanish.
\end{itemize}
After the pruning, the amount of residual elements is reduced by 64.3\% and the number of variables is reduced by 50\%, resulting in a speedup of 4 times for the optimization.

On a DJI Manifold 2-G on-board computer, in a scenario of aerial swarm with 5 drones, the average optimization time is reduced from 686.6ms to 169.6ms in a indoor flight case. Consider the drift of VIO can be negligible during such small peiod, our two-stage fusion algorithm can be treated as real-time.

\subsubsection{Estimator Initialization}\label{init}
Since the optimization problem of estimating the relative pose mentioned in Sec. \ref{optimization} is highly nonlinear, directly fusing the measurements summarized Sec. \ref{KFs} without an appropriate initial value is almost impossible. Besides, due to the lack of observability, the relative yaw rotation cannot be solved unless the drones have performed enough translational movements. Hence in practice, the assumption that the observed drones are facing to close directions is made during the initialization, which means the yaw angle differences are considered to be small.

% When initial we fixed $z_{i,k}^t$ and  $\psi_{i,k}^t$ of $\mathbf{P}_{i,k}^t$ to $z_{i}^t$ and  $\psi_{i}^t$, only solving $x_{i,k}^t$ and $y_{i,k}^t$ on the sliding window.
% $$
% \left\{
%     \begin{array}{ll}
%       \max_t x_{k}^t - \min_t x_{k}^t > 2.0\\
%       \max_t y_{k}^t - \min_t y_{k}^t > 2.0 \\
%       \max_t z_{k}^t - \min_t z_{k}^t > 1.0
%     \end{array}
%   \right.
% $$
  
For drone \textit{k} with sufficient movement, a few times of trial of optimization with random initial translational values are performed and the result with the smallest cost will be selected as the initialization result. Once the initialization has finished, the normal optimization will proceed with the obtained initial value.

\section{Detailed System Implementation}\label{swarmsys}

In order to verify the validity of the relative state estimation framework stated in Sec.\ref{rel_est}, a customized aerial swarm system consists of aerial platforms and a user interface (UI) is designed and implemented. All the state estimation and planning algorithms are running on the individual aerial platforms, while the ground station is only utilized for sending commands and monitoring flight status. The entire system is built under the ROS framework.

\subsection{Aerial Platform}\label{thedrone}
\begin{figure}[ht!]
\centering
% \begin{subfigure}{\textwidth}
\centering
\includegraphics[width=0.9\linewidth]{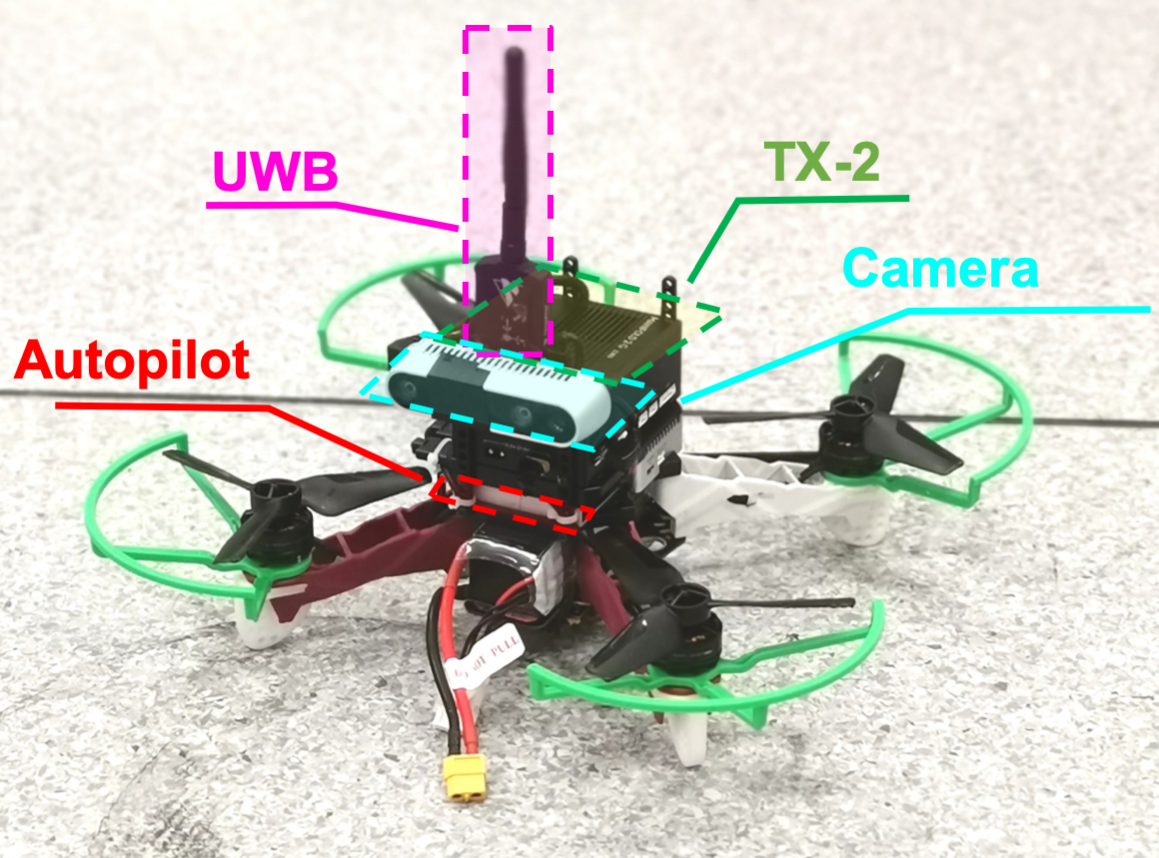}
\vspace{-0.1cm}
%\caption{UAV}
% \small
% The drone in aerial swarm
\caption{One of the aerial platforms in the swarm system, which is equipped with a RealSense D435i depth camera, a DJI N3 flight controller, a Nooploop UWB module and a DJI Manifold2-G on-board computer with a Nvidia TX-2 chip.}\label{drone}
\vspace{-0.7cm}
\end{figure}

Five aerial platforms with identical hardware configurations are adopted for the swarm system. One of the platforms, which is a quad-rotor drone, is shown in Fig.\ref{drone}. The drone is equipped with a DJI N3 flight controller with an IMU embedded in, an Intel RealSense D435i stereo camera module, a Nooploop UWB module and a DJI Manifold2-G on-board computer. The raw data from the IMU at 400Hz and images from the stereo camera at 20Hz are fused together for the visual-inertial state estimation of individual drones using the GPU accelerated VINS-Fusion\cite{qin2019a}, while the images are also utilized for relative state estimation together with the distance measurements from the UWB module. The attitude control of the drone is performed by the flight controller, while all other computations, including position control, state estimation and trajectory planning, are executed on the on-board computer. The communications between the drones as well as between the drones and the ground station are through the UWB modules, which are broadcasting data at 100Hz with synchronized time stamps. 
\vspace{-0.3cm}

% \subsection{User Interface}\label{ground_station}
% For the purpose of operational convenience, a WebGL based user interface (UI) is developed, as shown in Fig.\ref{gcs}. The interface displays position, battery level, control mode and all other useful real-time information received from the drones that can indicate the flight status. Besides, the absolute and relative positions of the aerial platforms are shown graphically using their 3-D models. Furthermore, the UI is designed to be interactive. In addition to the traditional auto takeoff and landing buttons, users can drag the selected drones to a desired position and send the command back. Since the ground station system is designed for swarm flights, there are buttons for sending commands of changing formations as well.

% \begin{figure}[h!]
% \centering
% \vspace{-0.2cm}
% \includegraphics[width= 1.0\linewidth]{gcs}
% \vspace{-0.4cm}
% \caption{A snapshot of the user interface during a swarm formation flight. The real-time flight statuses are shown and the commands can be sent through the interacting interface.}\label{gcs}
% \vspace{-0.6cm}
% \end{figure}

\begin{figure*}[t]
\centering
\subfigure[\label{01_result} Relative positions of Drone 0\&1]
{\includegraphics[width=0.24\linewidth]{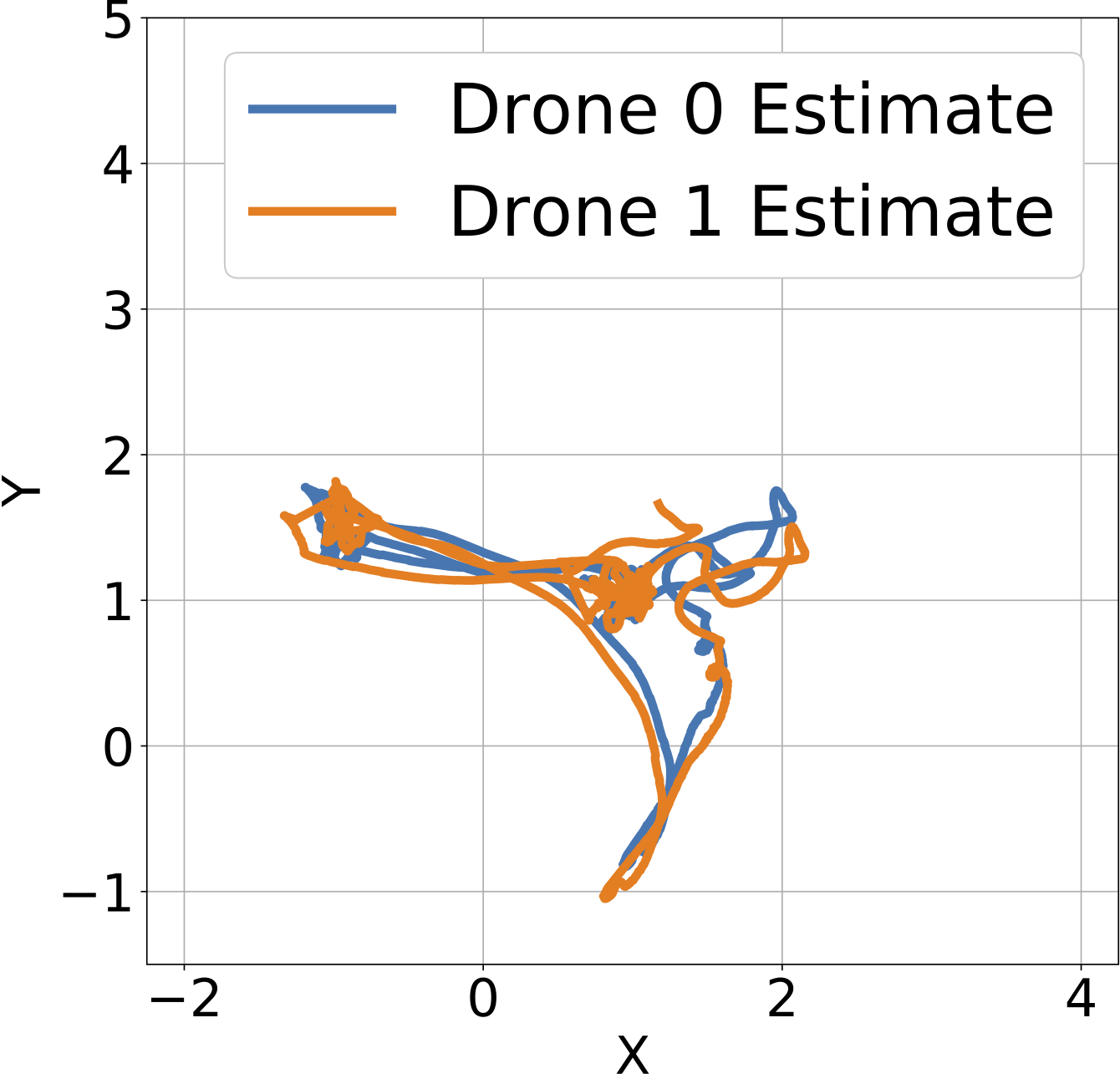}}
\subfigure[\label{02_result} Relative positions of Drone 0\&2]
{\includegraphics[width=0.24\linewidth]{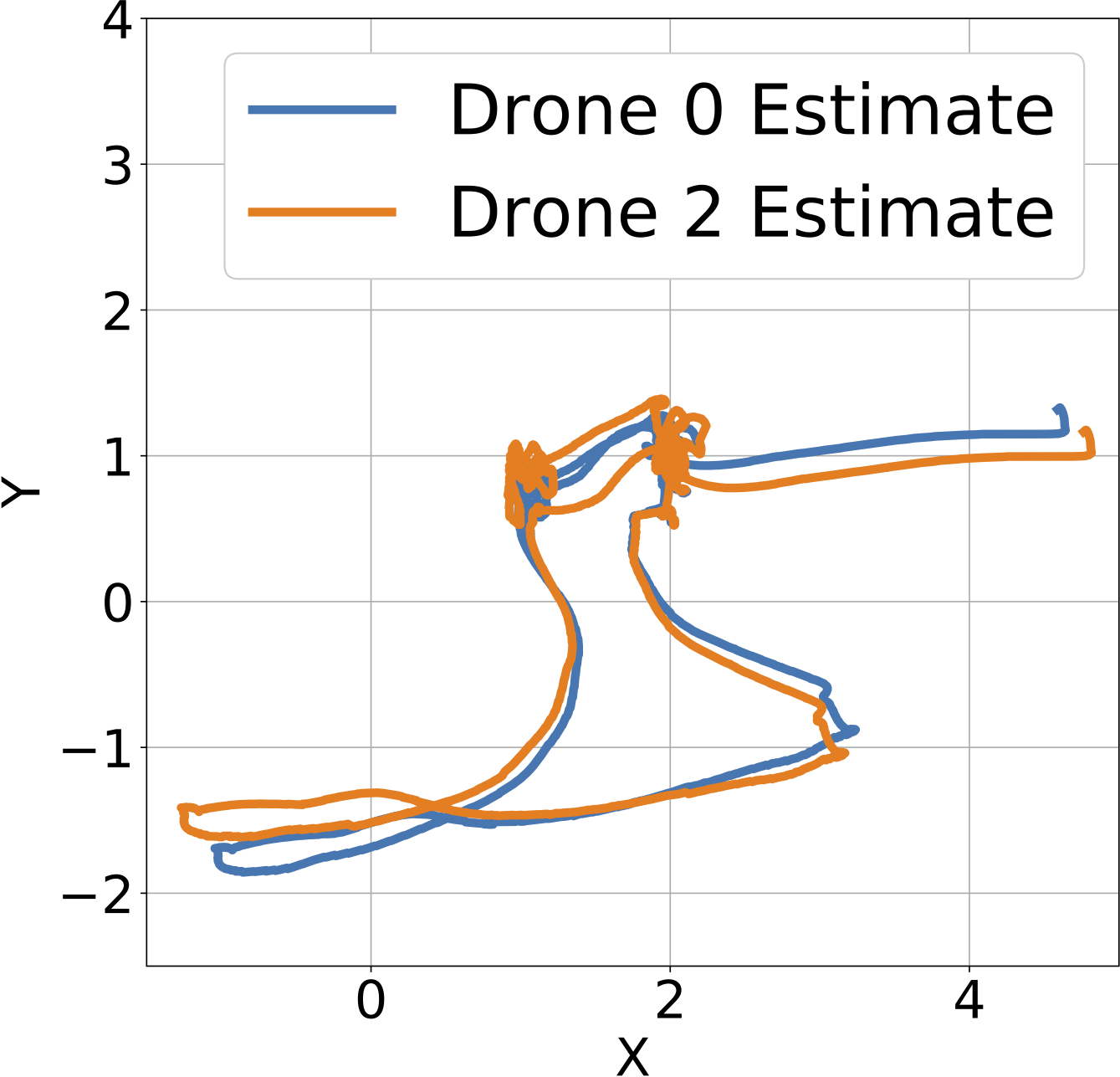}}
\subfigure[\label{03_result} Relative positions of Drone 0\&3]
{\includegraphics[width=0.24\linewidth]{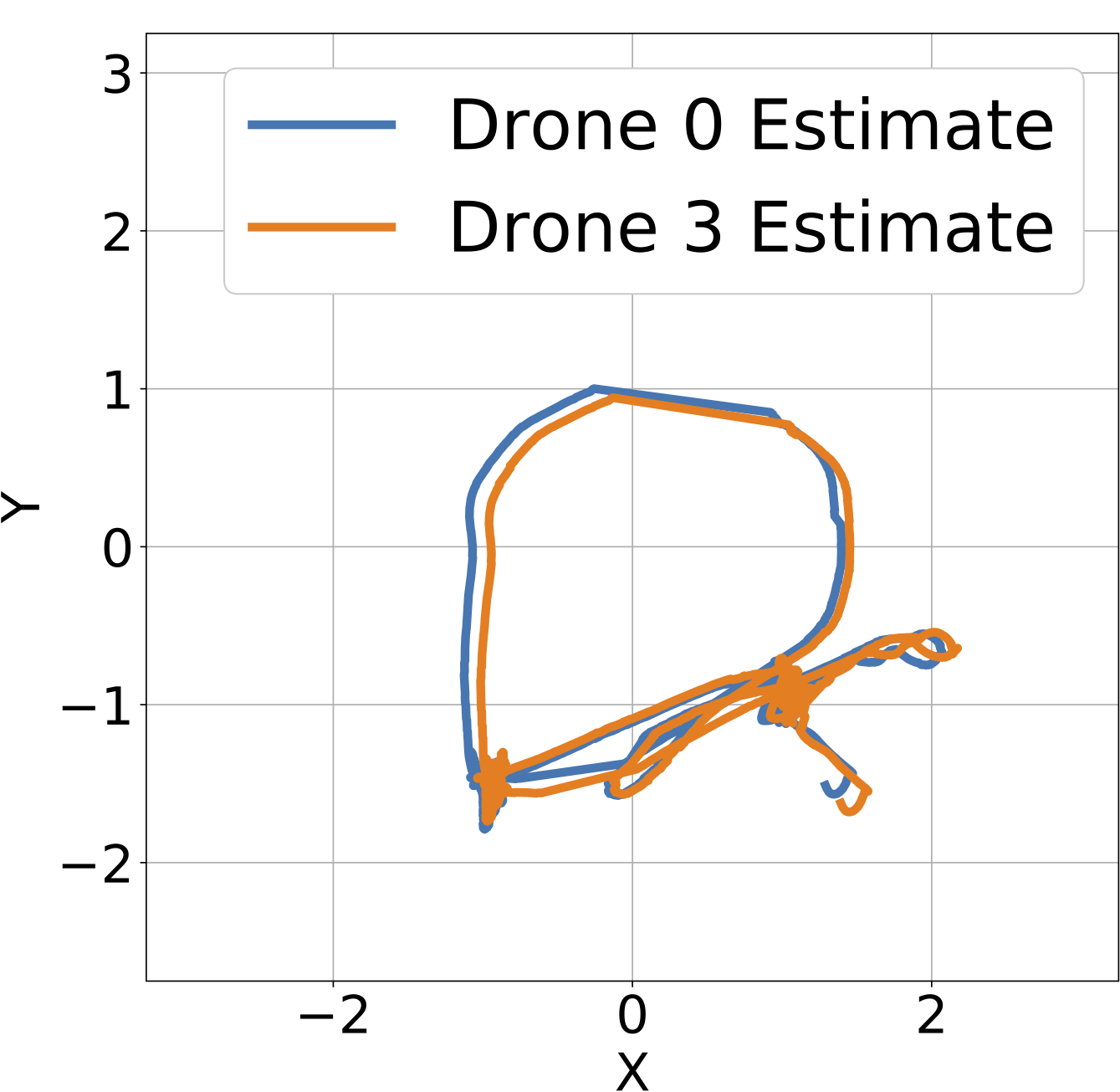}}
\subfigure[\label{04_result} Relative positions of Drone 0\&4]
{\includegraphics[width=0.24\linewidth]{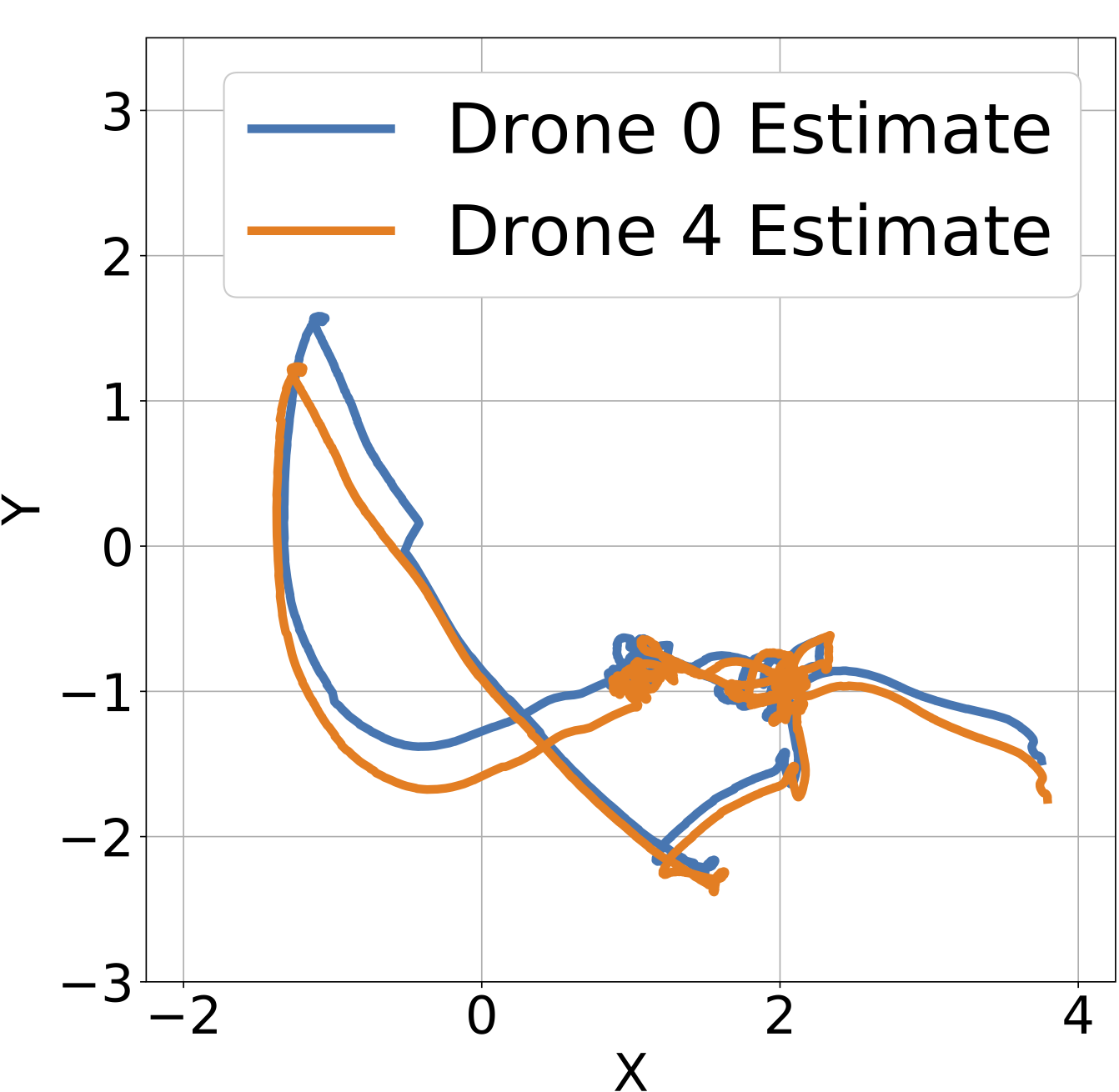}}
\vspace{-0.2cm}
\caption{The comparison of the real-time relative estimation between drone 0 and other drones in the formation flight with 5 drones. The relative estimations of drone 0 from other drone are inverted for comparison. 
\label{5_drone_result}}\label{5drones}
\vspace{-0.7cm}
\end{figure*}

\section{Experiment and Result}
\subsection{Experiment Setup}
The experiments are conducted in two indoor scenarios with the aerial swarm system mentioned in Sec. \ref{swarmsys}. The system consists of up to 5 drones is tested in two different indoor scenarios and compared with the ground truth data obtained from the OptiTrack motion capture system. A trajectory planner based on our estimated state for formation flying is introduced in the experiments to verify the validity and effectiveness of our relative state estimation method. The planner supports the transformation between 5 formations as well as translations of the swarm system. The traditional minimum-snap trajectory generator from way-points~\cite{mellinger2011minimum} is adopted, and the way-points are preset to avoid collisions during the changes of the formations.
\subsection{Experiment Result and Comparison}\label{comparison}
The relative state estimation framework is tested with 2, 3 and 5 aerial vehicles. Motion capture System is introduced as ground truth to ensure accuracy by comparing the relative poses from the motion capture system and our method.

\begin{figure}[t]
    \centering
    \includegraphics[width=0.8\linewidth]{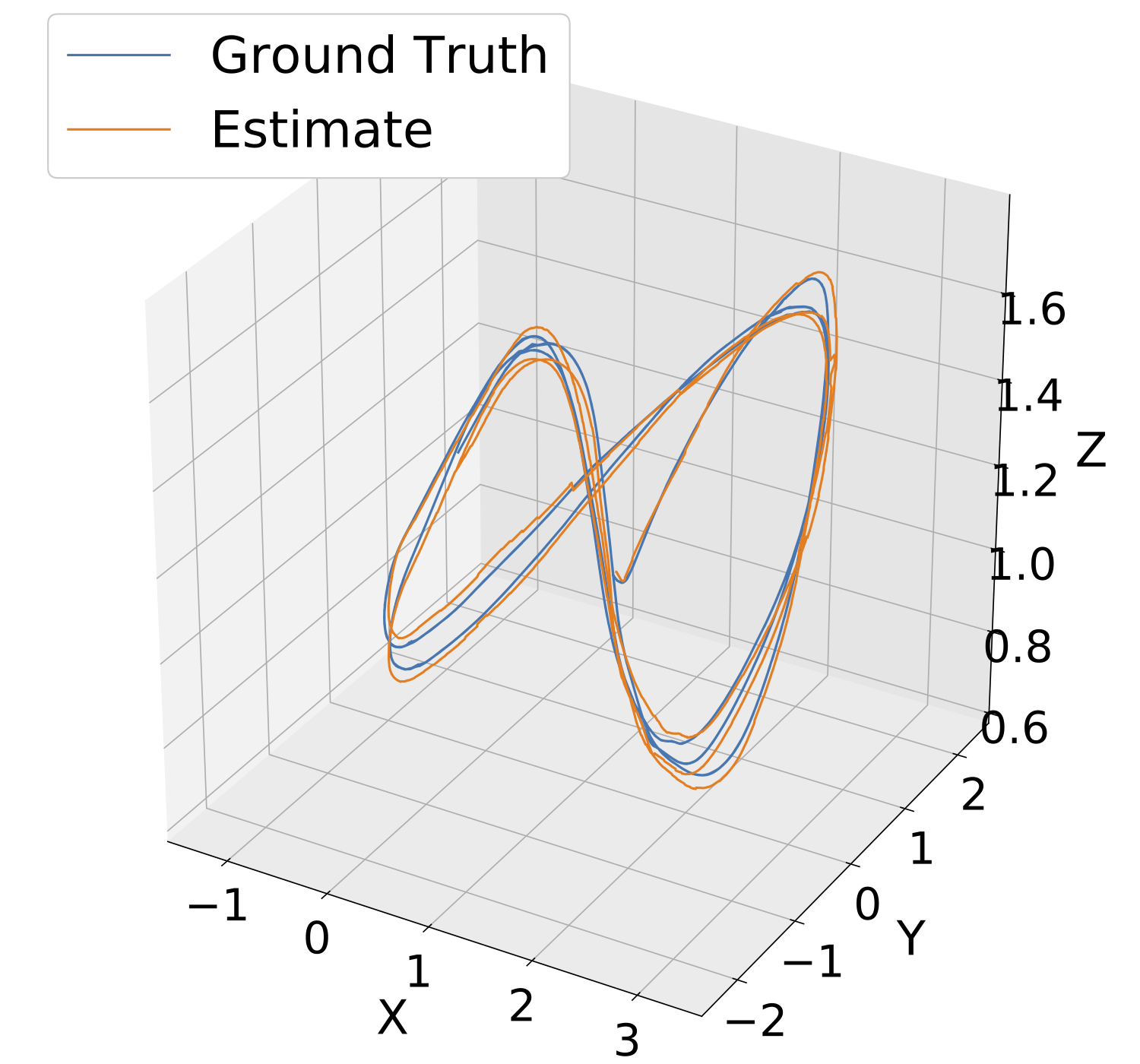}
    \vspace{-0.2cm}
    \caption{The comparison of the observed trajectory of a drone between the relative estimation and the ground truth in two-drone scenario. The two frames are aligned for the visualization purpose. The RMSE of translation and yaw rotation are 6.84cm and 2.131 degree during the 34.553m flight.}
    \label{fig:error}
    \vspace{-0.7cm}
\end{figure}

Fig. \ref{fig:error} shows the estimated trajectory and the trajectory recorded by the motion capture system in a two-drone scenario while Tab. \ref{tab:vicon_compare1} and Tab. \ref{tab:vicon_compare2} show the error comparison between the proposed method and the ground truth. The UWB data and the detection data are abandoned in different scenarios to verify the effectiveness of the proposed sensor fusion method. According to Tab. \ref{tab:vicon_compare1} and \ref{tab:vicon_compare2}, it is shown that the RMSE error can reach 5cm level in 2 drones' aggressive flight and is similar with a triple-drone system.
It should be noted that without the UWB measurements, it is extremely difficult for the system to initialize with pure visual detection because of the difficulty in depth estimation and making distinctions between individual drones. Hence the benchmarked result without UWB is utilizing the initialization of the full system. For the five drones scenario, due to the experiment site limitation, the real-time relative state between drone \textit{0} and other drones are provided in Fig. \ref{5_drone_result} to show the consistency of our proposed method.

% Due to our detection algorithm, the result of visual only is first initialize with UWB measurement, after system is initialized, we disable all UWB residuals to show the performance of visual only method with good initialization and matching.

% The results shows that for our proposed method, the system gives best result and the real-time estimated trajectory is smooth. When using UWB measurement only, the error compare to motion capture will be larger. When using detection the RMSE is slightly bigger than our proposed method on x, y axis, which shows when using a good initialization from UWB measurement, our method gives an also good result. However, it's not possible to use visual only, because with visual only it's hard to distinguish drones, and when running on larger scale(more than 3m), detection will become very hard on onboard computer.

When comparing with previous UWB based methods, the estimation system with ground anchors proposed in\cite{ledergerber2015robot} reaches an RMSE error of 0.14m on the horizontal position and 0.28m on the total error, while \cite{guo2019ultra} mentioned in Sec. \ref{related_works} provides an error of 0.2m horizontally, without mentioning the vertical error, both of which are outperformed by the proposed decentralized relative state estimation, in terms of either completeness or precision. On the other hand, compared with the vision based system in\cite{saska2016auro}, which provides 10 to 20cm estimation error, the proposed method provides better precision, meanwhile has the ability of handling the situation when the other vehicles are out of the FoV. 

Furthermore, the consistency demonstrated in Fig. \ref{5_drone_result} shows the convenience of applying the method in different scales. Furthermore, the fully decentralized visual-inertial-UWB system is not limited by the FoV or GPS, which means that it has the potential to be prevalently adopted by aerial swarm applications in different environments.
{\small
\begin{table}[t]
\begin{center}
\caption{Comparison of Relative State Estimation with Ground Truth with a Double-Drone Setting}
\label{tab:vicon_compare1}
\begin{tabular}{|c|c|c|c|}
\hline
 & \multirow{2}{*}{\begin{tabular}{@{}c@{}}\textbf{Proposed} \\ \textbf{Method}\end{tabular}} 
 & \multirow{2}{*}{\begin{tabular}{@{}c@{}}Without \\ UWB \end{tabular}} 
 & \multirow{2}{*}{\begin{tabular}{@{}c@{}}Without \\ Detection \end{tabular}} 
\\ &&& \\
\hline
Traj. Length  & \multicolumn{3}{c|}{34.553m} \\
\cline{1-4}
RMSE of $^{b_k}x_{i}^t$ &\textbf{0.029m} &0.063m  &0.038m\\
\cline{1-4}
RMSE of $^{b_k}x_{i}^t$ &\textbf{0.058m} &0.076m  &0.224m\\
\cline{1-4}
RMSE of $^{b_k}z_{i}^t$ &\textbf{0.022m} &0.018m  &0.046m \\
\cline{1-4}
RMSE of $^{b_k}\psi_{i}^t$ &\textbf{2.131 deg} &3.665 deg &3.488 deg \\
\cline{1-4}
\end{tabular}
\end{center}
\vspace{-0.3cm}
\end{table}

\begin{table}[t]
\begin{center}
\caption{Comparison of Relative State Estimation with Ground Truth with a Triple-Drone Setting}
\label{tab:vicon_compare2}
\begin{tabular}{|c|c|c|c|}
\hline
& \multirow{2}{*}{\begin{tabular}{@{}c@{}}\textbf{Proposed} \\ \textbf{Method}\end{tabular}} 
& \multirow{2}{*}{\begin{tabular}{@{}c@{}}Without \\ UWB \end{tabular}} 
& \multirow{2}{*}{\begin{tabular}{@{}c@{}}Without \\ Detection \end{tabular}} 
\\ &&& \\
\hline
Traj. Length  & \multicolumn{3}{c|}{16.382m} \\
\cline{1-4}
RMSE of $^{b_k}x_{i}^t$ &\textbf{0.046m} &0.061m & 0.072m \\
\cline{1-4}
RMSE of $^{b_k}x_{i}^t$ &\textbf{0.052m} &0.075m &0.069m \\
\cline{1-4}
RMSE of $^{b_k}z_{i}^t$ &\textbf{0.013m} &0.031m &0.176m \\
\cline{1-4}
RMSE of $^{b_k}\psi_{i}^t$ &\textbf{1.015 deg} &2.106 deg & 3.070 deg\\
\cline{1-4}
\end{tabular}
\end{center}
\vspace{-0.5cm}
\end{table}
}

\section{Conclusion and Future Work}
\vspace{-0.2cm}

In this paper, a novel decentralized visual-inertial-UWB relative state estimation framework is introduced. The proposed method combines detection, VIO and distance measurements together and provide the estimation result based on the optimization. The relative estimation framework is tested by extensive aerial swarm flight experiments with up to 5 customized drones and a ground station to demonstrate the feasibility and effectiveness. The relative estimation results are compared with ground truth data from the motion capture system, reaching a precision of centimeter-level, which outperforms all the Ultra-WideBand and vision based methods until present. On account of the flexibility and reliability of the system, this is the first fully decentralized relative state estimation framework that can be taken into practical applications. We believe that the proposed framework has the potential to be widely adopted by aerial swarms in different scenarios and in multiple scales.

With the proposed estimator, formation flights in various complex environments are no longer impossible. In the future, different control and planning algorithms for aerial swarm can be developed based on the relative estimation framework. We are planning to expand the system to be capable of performing safe flights in cluttered environments and even collaborative search \& rescue tasks.

%\newlength{\bibitemsep}\setlength{\bibitemsep}{.03\baselineskip}
%\newlength{\bibparskip}\setlength{\bibparskip}{0pt}
%\let\oldthebibliography\thebibliography
%\renewcommand\thebibliography[1]{%
%  \oldthebibliography{#1}%
%  \setlength{\parskip}{\bibitemsep}%
%  \setlength{\itemsep}{\bibparskip}%
%}
\bibliographystyle{IEEEtran}
\bibliography{ICRA2020hao} 

\end{document}